\pdfoutput=1

\documentclass[11pt]{article}

\usepackage[preprint]{acl}

\usepackage{times}
\usepackage{latexsym}

\usepackage[T1]{fontenc}

\usepackage[utf8]{inputenc}
\DeclareUnicodeCharacter{266B}{\textmusicalnote}
\usepackage{microtype}

\usepackage{inconsolata}

\usepackage{graphicx}

\usepackage{booktabs}
\usepackage{multirow}
\usepackage{booktabs}
\usepackage{graphicx}
\usepackage{url}

\title{MRCEval: A Comprehensive, Challenging and Accessible Machine Reading Comprehension Benchmark}

\author{Shengkun Ma, Hao Peng, Lei Hou, Juanzi Li \\
  Department of Computer Science and Technology, Tsinghua University \\
  \texttt{mashengkun98@gmail.com}}

\begin{document}
\maketitle
\begin{abstract}
Machine Reading Comprehension (MRC) is an essential task in evaluating natural language understanding. Existing MRC datasets primarily assess specific aspects of reading comprehension (RC), lacking a comprehensive MRC benchmark. To fill this gap, we first introduce a novel taxonomy that categorizes the key capabilities required for RC. Based on this taxonomy, we construct \textbf{MRCEval}, an MRC benchmark that leverages advanced Large Language Models (LLMs) as both sample generators and selection judges. MRCEval is a comprehensive, challenging and accessible benchmark designed to assess the RC capabilities of LLMs thoroughly, covering 13 distinct RC skills with a total of 2.1K high-quality multi-choice questions. We perform an extensive evaluation of 28 widely used open-source and proprietary models, highlighting that MRC continues to present significant challenges even in the era of LLMs. Project is available at github. \footnote{\url{https://github.com/THU-KEG/MRCEval}}
\end{abstract}

\section{Introduction}
With the advancement of Large Language Models (LLMs), such as o3-mini \citep{o3-mini} and DeepSeek-R1 \citep{guo2025deepseek}, their remarkable language understanding and generation capabilities continue to impress AI communication. Machine Reading Comprehension (MRC), which requires machine reading and comprehending the given passage, then answering the questions correctly, is the fundamental evaluation of natural language understanding \citep{hirschman-etal-1999-deep}.

\begin{figure}
\includegraphics[scale=0.45]{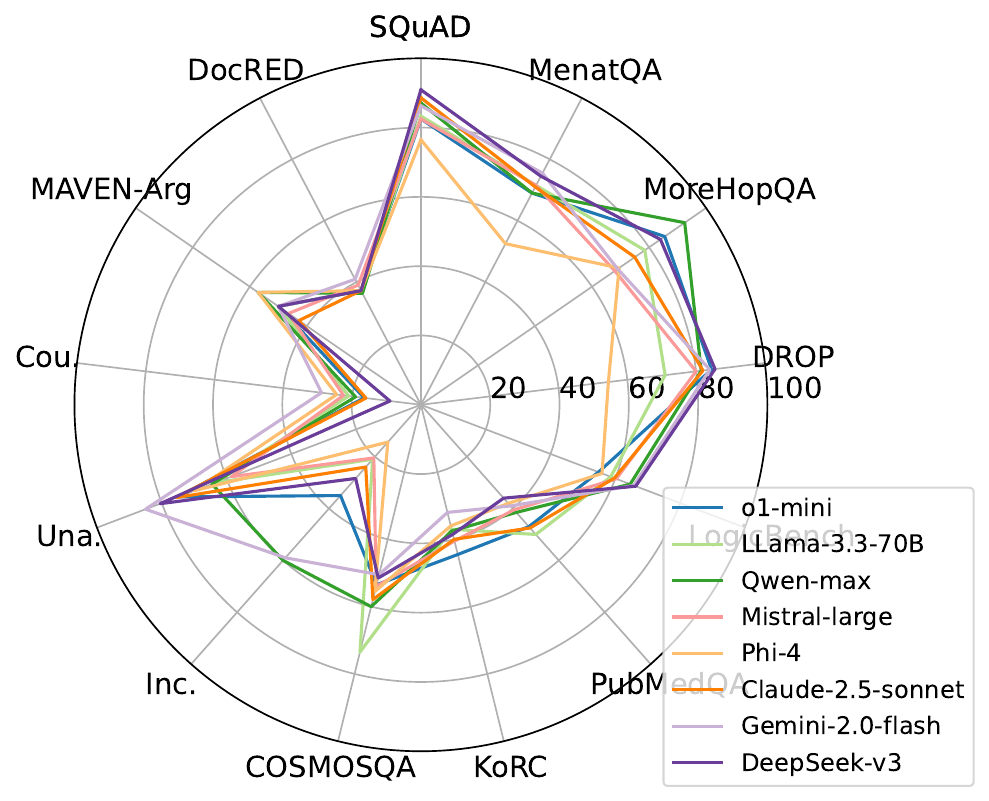}
\caption{Performance on MRCEval Benchmark of representative models.}
\label{figure:teaser}
\centering
\end{figure}

To facilitate the reading comprehension (RC) capability of the machine, a great number of datasets are proposed \citep{rajpurkar-etal-2016-squad, yang-etal-2018-hotpotqa, trivedi-etal-2022-musique, yao-etal-2023-korc, parmar-etal-2024-logicbench}. However, all these datasets only focus on a specific RC skill, and there is a lack of a unified benchmark to evaluate the RC challenges of LLMs. Current MRC taxonomies are fine-grained, complicated, and not suitable for creating a comprehensive but accessible MRC benchmark. On the other hand, there are many new issues appeared with LLMs, such as hallucination \citep{ji2023survey} and knowledge conflict \citep{xu2024knowledge}. These issues make LLMs unable to accurately understand the given context and answer the question incorrectly, making RC more challenging for LLMs.

To address these issues, we first introduce a novel taxonomy for MRC. Drawing inspiration from the machine's question-answering process of text comprehension \citep{lehnert1978process}, we summarize the needed RC skills for LLMs into three levels: context comprehension, external knowledge comprehension and reasoning. As \citet{mccarthy1990example} said, machines first understand the facts in the passage, then grasp the expression of the general information about the world that could allow getting the answers to the questions by formal reasoning from the facts and the general information. These levels correspond to the accurate comprehension of facts information, the acquisition of external knowledge, and the integration of facts and expertise for reasoning.

Based on the proposed taxonomy, we introduce MRCEval, a comprehensive, challenging and accessible MRC benchmark designed to assess RC capabilities of LLMs. MRCEval comprises three main tasks with 13 sub-tasks, and a total of 2.1K high-quality multi-choice questions. It is built employing GPT-4o \citep{hurst2024gpt} as the generator, and three light-weight models as judges to generate the high-quality and challenging samples.

We conduct an extensive evaluation on 28 representative open-source and closed-source models. Figure \ref{figure:teaser} summarizes the performance of popular competitive models on the 13 sub-tasks in MRCEval. It reveals that MRC still remains challenging, even the most competitive models like o1-mini and Gemini-2.0-flash still perform badly on MRCEval, despite their strong performance on standard benchmarks.  As far as we know, MRCEval is the first comprehensive, challenging and accessible benchmark tailored for MRC, contributing to the advancement of natural language understanding in LLMs.

\section{Related Work}
\paragraph{MRC datasets.}
Numerous datasets have been proposed in the past decade. Cloze-test format (SQuAD 1.0 \citep{rajpurkar-etal-2016-squad}), free-answer format (NarrativeQA \citep{kocisky-etal-2018-narrativeqa}), arithmetic (DROP \citep{dua-etal-2019-drop}), commonsense (OpenBookQA \citep{mihaylov-etal-2018-suit}), world knowledge (Natural Questions \citep{kwiatkowski-etal-2019-natural}), reasoning (HotpotQA \citep{yang-etal-2018-hotpotqa}), logical reasoning (ReClor \citep{yu2020reclor}), multi-hop reasoning (MorehopQA \citep{schnitzler2024morehopqa}), temporal reasoning (TORQUE \citep{ning-etal-2020-torque}), medical (MedMCQA \citep{pal2022medmcqa}) and science (ScienceQA \citep{lu2022learn}).
\paragraph{Benchmarking MRC.}
Researchers summarize RC skills in different aspects. \citet{chen2018neural} first defines the MRC task depending on the answer type: cloze style, multiple choice, span prediction, and free-form answer. Then \citet{schlegel-etal-2020-framework} analyze modern MRC gold standards and propose a qualitative annotation schema to evaluate popular MRC datasets. \citet{sugawara-etal-2021-benchmarking} further provide a theoretical basis for the design of MRC datasets based on psychology as well as psychometrics and summarize it in terms of the prerequisites for benchmarking MRC. Based on these research, \citet{rogers2023qa} propose an alternative taxonomy for a wider range of RC skills.

\section{MRCEval Benchmark}

\subsection{Taxonomy}
Building on the three levels of machine text comprehension \citep{mccarthy1990example}, we define three key aspects of MRC: \textbf{Context Comprehension}, \textbf{External Knowledge Comprehension}, and \textbf{Reasoning}.

\paragraph{Context comprehension.} Focuses on facts understanding and models' context-faithful capability \citep{ming2024faitheval}. First, models should understand the facts, which include entities, relations and events related facts in the text. Then overcome the hallucination from their parameters to be faithful to the given context.
\paragraph{Externation knowledge comprehension.} Focuses on external knowledge acquisition and application \citep{wang2021cognet}. Models are supposed to incorporate the external knowledge outside the given text, which is from the real world, namely the world general knowledge, commonsense knowledge, and the specific domain knowledge to comprehend the passage and the question.
\paragraph{Reasoning.} Focuses on deep context comprehension and inference, which is an essential capability for complex problem-solving \citep{qiao-etal-2023-reasoning}. In the MRC task, we classify reasoning into logical reasoning, arithmetic reasoning, multi-hop reasoning and temporal reasoning.

\begin{table}
\centering
\resizebox{0.45\textwidth}{!}{
\begin{tabular}{lcc}
\toprule
\textbf{Task} & \textbf{Dataset} & \textbf{Instances} \\
\midrule
\textbf{Context comprehension} & - & \textbf{774} \\
\midrule
Facts understanding &  &  \\
-entity & SQuAD 1.0 & 132 \\
-relation & DocRED & 24 \\
-event & MAVEN-Arg & 18 \\
Context faithful &  &  \\
-counterfactual & FaithEval & 200 \\
-unanswerable & FaithEval & 200 \\
-inconsistent & FaithEval & 200 \\
\midrule
\textbf{External knowledge comprehension} & - & \textbf{545} \\
\midrule
Commonsense knowledge & COSMOSQA & 200 \\
World knowledge & KoRC & 145 \\
Domain knowledge & PubMedQA & 200 \\
\midrule
\textbf{Reasoning} & - & \textbf{784} \\
\midrule
Logical reasoning & LogicBench & 184 \\
Arithmetic reasoning & DROP & 200 \\
Multi-hop reasoning & MoreHopQA & 200 \\
Temporal reasoning & MenatQA & 200 \\
\midrule
\textbf{Overall} &  & \textbf{2103} \\
\bottomrule
\end{tabular}}
\caption{MRCEval tasks division.}
\label{table:division}
\end{table}

\subsection{Benchmark Construction}
Based on the proposed taxonomy, we construct the MRCEval benchmark.
\paragraph{Source datasets.} 
For each sub-task, we collect representative datasets, including SQuAD 1.0 \citep{rajpurkar-etal-2016-squad}, DocRED \citep{yao-etal-2019-docred}, and MAVEN-Arg \citep{wang-etal-2024-maven}, FaithEval \citep{ming2024faitheval}, COSMOSQA \citep{huang-etal-2019-cosmos}, KoRC \citep{yao-etal-2023-korc}, PubMedQA \citep{jin-etal-2019-pubmedqa}, LogicBench \citep{parmar-etal-2024-logicbench}, DROP \citep{dua-etal-2019-drop}, MoreHopQA \citep{schnitzler2024morehopqa} and MenatQA \citep{wei-etal-2023-menatqa}. Due to access rights, we use their development sets.

\paragraph{Multi-choice samples construction.}
For multi-choice format datasets COSMOSQA, LogicBench, and counterfactual part of FaithEval, we have retained the original data. For some question-answering datasets SQuAD 1.0, KoRC, DROP, MenatQA, and unanswerable part of FaithEval, we set the answer as the correct option and prompt GPT-4o to generate three incorrect options. For others, we use the automated method to construct three incorrect options. As for DocRED and MAVEN-Arg, since they have no questions, we prompt GPT-4o to generate a facts-related question and options for each passage as an instance. 
\paragraph{LLMs as judges.}
To select challenging instances for LLMs, we adopt a voting strategy employing three light-weight LLMs as judges: LLama-3-8B-Instruct \citep{dubey2024llama}, Qwen-2.5-7B-Instruct \citep{yang2024qwen2}, and GPT-4o-mini \citep{hurst2024gpt}. For each sample, if at least one of the judges answers incorrectly, we put the sample as the candidate. After noise filtering, we randomly select some of the remaining candidates for each sub-task to build the final benchmark.
\paragraph{Statistic.}
MRCEval is an English benchmark, which consists of general topics with three main tasks: context comprehension, external knowledge comprehension, reasoning, and 13 sub-tasks. As Table \ref{table:division}, context comprehension includes facts understanding (entity, relation and event facts) and context-faithful (counterfactual, unanswerable and inconsistent). External knowledge comprehension includes world knowledge, commonsense knowledge and domain knowledge. Reasoning includes logical reasoning, arithmetic reasoning, multi-hop reasoning and temporal reasoning. In sum, MRCEval has 2,103 multi-choice instances.

\section{Evaluation}
\paragraph{Models.}
We evaluate extensive popular open-source and closed-source models. For open-source models, we consider their instruction-tuned models, including LLama-3.1-8B-Instruct, LLama-3.3-70B-Instruct \citep{dubey2024llama}, Qwen-2.5-14B-Instruct \citep{yang2024qwen2}, Mistral-7B-Instruct-v0.3, Mistral-Nemo-Instruct-2407, Mistral-8x7B-Instruct-v0.1 \citep{jiang2023mistral}, Gemma-2-9B-it, Gemma-2-27B-it \cite{team2024gemma}, Phi-3-mini-4k-Instruct, Phi-3-medium-4k-Instruct, Phi-4 \citep{abdin2024phi}, Command-R-7B-12-2024 \citep{command-r}, DeepSeek-R1-Distll-LLama-8B, DeepSeek-R1-Distll-Qwen-14B \citep{guo2025deepseek}, DeepSeek-v3 \citep{liu2024deepseek}. For closed-source models, we access them through their official API, including GPT-3.5-turbo, GPT-4-turbo, GPT-4o, o1-mini, o3-mini, Gemini-1.5-flash, Gemini-1.5-pro, Gemini-2.0-flash, Gemini-2.0-flash-lite-preview-02-05, Claude-3.5-haiku-20241022, Claude-3.5-sonnet-20241022, Mistral-large and Qwen-max-2025-01-25.
\paragraph{Settings.}
All models use greedy sampling or temperature of 0.0, except for DeepSeek series, which follows their official settings with temperature of 0.60 and top-p of 0.95.
For all tasks, we append the instruction to the beginning of each instance: \textit{You are an expert in reading comprehension. Read the passage and select one of the most appropriate options to answer the question.} We report accuracy as the metric from a single run result.

\section{Results and Analysis}

\begin{table*}
\centering
\resizebox{\textwidth}{!}{
\begin{tabular}{lccccccccccccccccc}
\toprule
\multirow{3}{*}{\textbf{Models}} & \multicolumn{6}{c}{\textbf{Context}} & \multicolumn{3}{c}{\textbf{External Knowledge}} & \multicolumn{4}{c}{\textbf{Reasoning}} & \multicolumn{4}{c}{\textbf{Overall}} \\
\cmidrule(l{10pt}r{10pt}){2-7} \cmidrule(l{10pt}r{10pt}){8-10} \cmidrule(l{10pt}r{10pt}){11-14} \cmidrule(l{10pt}r{10pt}){15-18}
& \multicolumn{3}{c}{\textit{Facts Understanding}}  & \multicolumn{3}{c}{\textit{Context Faithful}} & \multirow{2}{*}{Com.} & \multirow{2}{*}{Wor.} & \multirow{2}{*}{Dom.} & \multirow{2}{*}{Log.} & \multirow{2}{*}{Ari.} & \multirow{2}{*}{Mul.} & \multirow{2}{*}{Tem.} & \multirow{2}{*}{Con.} & \multirow{2}{*}{Kno.} & \multirow{2}{*}{Rea.} & \multirow{2}{*}{\textbf{Avg.}} \\
& Ent. & Rel. & Eve. & Cou. & Una. & Inc. \\
\midrule
\multicolumn{18}{c}{Open-source Models} \\
\midrule
Mistral-7B-Instruct-v0.3 & 57.5 & 30.9 & 50.0 & \textbf{52.0} & 27.5 & 3.0 & 45.5 & 30.5 & 36.5 & 36.9 & 31.5 & 22.5 & 31.5 & 33.2 & 37.5 & 30.4 & 33.4 \\
Mistral-8x7B-Instruct-v0.1 & 59.0 & 31.8 & \textbf{57.1} & 45.0 & 39.5 & 8.5 & 54.0 & 28.0 & 39.5 & 46.1 & 41.5 & 31.5 & 43.0 & 36.2 & 40.5 & 40.4 & 38.8 \\
Mistral-Nemo-Instruct-2407 & 68.1 & 33.6 & 39.2 & 39.5 & 42.5 & 5.5 & 48.5 & 32.0 & 40.0 & 44.0 & 37.0 & 31.0 & 57.9 & 35.9 & 40.1 & 42.4 & 39.3 \\
Phi-3-mini-4k-Instruct & 71.2 & 31.8 & \textbf{57.1} & 29.5 & 33.5 & 8.5 & 47.5 & 41.5 & 25.5 & 47.8 & 43.0 & 63.5 & 42.0 & 33.1 & 38.1 & 49.1 & 40.0 \\
Phi-3-medium-4k-Instruct & 76.5 & \textbf{40.0} & 46.4 & 13.5 & 45.0 & 9.5 & 54.5 & 24.5 & 31.5 & 43.4 & 62.0 & 53.0 & 45.5 & 33.7 & 36.8 & 51.1 & 40.6 \\
LLama-3.1-8B-Instruct & 62.1 & 31.8 & 46.4 & 47.0 & 38.0 & 33.5 & 59.0 & 37.0 & 38.5 & 42.4 & 34.5 & 37.0 & 43.0 & 42.2 & 44.8 & 39.2 & 41.8 \\
Gemma-2-9B-it & 77.2 & 32.7 & 39.2 & 32.0 & 56.9 & 16.0 & 53.5 & 35.5 & 30.5 & 51.6 & 52.5 & 50.0 & 52.5 & 41.2 & 39.8 & 51.6 & 44.4 \\
Phi4 & 76.5 & 37.2 & \textbf{57.1} & 24.5 & 70.0 & 14.4 & 55.5 & 36.0 & 38.0 & 55.9 & 54.5 & 69.5 & 52.5 & 43.2 & 43.1 & 58.1 & 48.4 \\
Command-R-7B-12-2024 & 79.5 & 35.4 & 53.5 & 43.0 & 70.0 & 37.0 & 35.0 & \textbf{43.5} & 38.5 & 61.9 & 46.0 & 34.0 & 68.0 & 52.7 & 39.0 & 52.2 & 48.9 \\
Gemma-2-27B-it & 82.5 & 36.3 & 35.7 & 30.5 & 61.5 & 26.0 & 53.0 & 39.0 & 36.5 & 47.8 & 50.0 & 69.5 & 63.5 & 45.4 & 42.8 & 57.9 & 49.0 \\
DeepSeek-R1-Distll-LLama-8B & 72.7 & 34.5 & \textbf{57.1} & 51.0 & 44.0 & 28.0 & 42.5 & 30.5 & 38.5 & 47.8 & 77.5 & 57.4 & 67.0 & 45.5 & 37.1 & 62.7 & 49.2 \\
Qwen-2.5-14B-Instruct & 84.8 & 33.6 & 50.0 & 24.5 & 76.5 & 9.5 & 58.0 & 35.5 & 34.0 & 58.7 & 67.0 & 70.0 & 66.0 & 44.1 & 42.5 & 65.6 & 51.2 \\
LLama-3.3-70B-Instruct & 83.3 & 37.3 & 50.0 & 21.0 & 63.5 & 21.0 & \textbf{73.5} & 37.0 & \textbf{50.0} & 58.7 & 71.0 & 78.5 & 71.5 & 43.2 & \textbf{53.5} & 70.2 & 55.3 \\
DeepSeek-R1-Distll-Qwen-14B & 79.5 & 35.4 & 46.4 & 36.0 & 69.5 & 25.0 & 45.0 & 35.0 & 40.0 & 53.8 & \textbf{85.5} & 77.5 & \textbf{74.5} & \textbf{48.0} & 40.0 & 73.2 & 54.6 \\
DeepSeek-v3 & \textbf{90.9} & 37.2 & 50.0 & 9.0 & \textbf{80.5} & \textbf{28.4} & 51.5 & 38.5 & 36.0 & \textbf{66.3} & \textbf{85.5} & \textbf{84.0} & \textbf{74.5} & 47.2 & 42.0 & \textbf{77.8} & \textbf{56.4} \\
\midrule
\multicolumn{18}{c}{Closed-source Models} \\
\midrule
Claude-3.5-haiku-20241022 & 75.7 & 36.3 & 53.5 & 25.0 & 48.5 & 8.0 & 51.0 & 31.0 & 34.0 & 47.8 & 43.5 & 30.0 & 44.5 & 36.5 & 38.6 & 41.3 & 38.7 \\
GPT-3.5-turbo & 67.4 & 31.8 & 50.0 & 12.5 & 59.0 & 21.5 & 50.0 & 34.0 & 30.0 & 47.3 & 41.5 & 46.0 & 48.5 & 37.2 & 38.0 & 45.8 & 40.4 \\
Gemini-1.5-flash & 84.0 & 34.5 & 46.4 & 28.9 & 60.0 & 9.0 & 53.0 & 33.5 & 47.0 & 59.2 & 62.0 & 78.0 & 69.0 & 41.1 & 44.5 & 67.2 & 51.1 \\
Gemini-2.0-flash-lite-preview-02-05 & 84.8 & 34.5 & 39.2 & \textbf{30.5} & 84.0 & 23.5 & 46.0 & 20.5 & 33.5 & 58.1 & 78.0 & 58.5 & 68.5 & 50.2 & 33.3 & 65.9 & 51.1 \\
Mistral-large & 82.5 & 39.0 & 46.4 & 22.5 & 58.5 & 20.5 & 54.0 & 40.5 & 40.5 & 60.3 & 80.0 & 68.0 & 71.5 & 42.2 & 45.0 & 70.1 & 52.7 \\
Gemini-1.5-pro & 88.6 & 36.3 & 42.8 & 23.5 & 66.5 & 9.5 & 56.4 & 43.5 & 39.0 & 65.2 & 81.0 & 89.0 & 69.5 & 42.2 & 46.3 & 76.4 & 55.2 \\
Claude-3.5-sonnet-20241022 & 88.6 & 37.2 & 42.8 & 16.0 & 75.5 & 24.0 & 57.9 & 40.0 & 47.5 & 59.7 & 82.0 & 75.0 & 71.0 & 46.0 & 48.5 & 72.1 & 55.8 \\
GPT-4o & \textbf{90.9} & 38.2 & 46.4 & 12.0 & 75.0 & 35.5 & 58.5 & 47.5 & 41.5 & 58.2 & 66.0 & 81.5 & 71.5 & 48.3 & 49.2 & 69.5 & 55.9 \\
GPT-4-turbo & 89.4 & 39.1 & 42.9 & 15.0 & 76.5 & 35.5 & 58.5 & \textbf{49.5} & 41.5 & 59.2 & 64.0 & 81.5 & 72.0 & 49.1 & 49.8 & 69.4 & 56.3 \\
o1-mini & 82.6 & 37.3 & 50.0 & 16.5 & 75.0 & 35.0 & 53.5 & 44.5 & 47.0 & 54.9 & 84.5 & 85.5 & 69.0 & 47.9 & 48.3 & 73.9 & 57.1 \\
o3-mini & 87.9 & 38.4 & 53.4 & 28.0 & 75.6 & 29.5 & 54.5 & 48.0 & \textbf{49.0} & 59.6 & \textbf{86.0} & 83.5 & 71.0 & 49.4 & \textbf{51.5} & 75.8 & 59.0 \\
Gemini-2.0-flash & 86.3 & \textbf{40.9} & 50.0 & 28.9 & \textbf{85.0} & 59.0 & 50.5 & 32.0 & 39.0 & \textbf{65.7} & 84.0 & 69.0 & \textbf{75.5} & \textbf{59.6} & 40.5 & 73.7 & 59.3 \\
Qwen-max-2025-01-25 & 87.1 & 36.3 & \textbf{57.1} & 19.0 & 64.5 & \textbf{60.0} & \textbf{60.0} & 37.5 & 41.5 & 64.6 & 81.5 & \textbf{92.5} & 69.0 & 52.6 & 46.3 & \textbf{77.1} & \textbf{59.4} \\
\bottomrule
\end{tabular}}
\caption{Accuracy (\%) of open-source and closed-source models in all tasks of MRCEval. The highest results are denoted in \textbf{bold} respectively.}
\end{table*}

\subsection{Overall Performance}
\paragraph{LLMs are good at facts extraction, while bad at context-faithful.}
LLMs have great performance at entity-facts understanding, which demonstrates that they can comprehend simple entity facts and are good at extracting entity answers directly from the passage. Stronger models can recognize which questions cannot be answered, they know to answer the question based on the given text, rather than their trained parameters. These two aspects confirm that LLMs have a good capability for simple information extraction. As for more complicated relation or event facts, models perform worse. Even the most competitive models, like Gemini-2.0-flash or Qwen-max, are still struggling with them. Large commercial models are better at inconsistent tasks but worse at counterfactual tasks, smaller open-source models do the opposite. This is because models with more parameters remember more facts and can easily fit them into memory, while smaller models are better at reasoning itself.
\paragraph{External knowledge still remains a challenge.}
Both large and small models perform almost equally poorly in commonsense knowledge and world knowledge comprehension, which indicates that increasing the parameter scales has little effect on the understanding and application of general knowledge. However, as for domain knowledge acquisition and application, LLama-3.3 with 70B parameters performs better than LLama-3.1 with 8B, which demonstrates that larger models have a stronger ability to learn new knowledge.
\paragraph{Large-scale models are good reasoners in MRC.}
Larger models perform well in reasoning tasks of MRC, even in complicated, more-hop reasoning tasks. Due to the recent research focus on model reasoning \cite{xu2025towards}, there has been a greater emphasis on reasoning when training large models, so LLMs are better at reasoning in reading comprehension than the other two aspects, especially the recently released reasoning models like o1-mini, o3-mini, and DeepSeek-R1 series.

\subsection{Error Analysis}
To assess which aspects are more challenging for all the LLMs, we collect the proportion of instances for each sub-task in which all the models predict incorrectly. We consider six well-performed models, LLama-3.3-70B-Instruct, Qwen-2.5-14B-Instruct, Gemini-2.0-flash, GPT-4o, o1-mini and Claude-3.5-sonnet-20241022. As Figure \ref{figure:error_study} shows, we find that all the models in counterfactual faithful task have the most common prediction errors, which indicates that all models overly rely on the memorization of parameters rather than truly understanding the text. While in other tasks, such as entity understanding and multi-hop reasoning, models have different agreements, meaning that these aspects are not common weaknesses of all the models.

\section{Conclusion}
In this work, we propose a novel MRC taxonomy and build a comprehensive, challenging and accessible MRC benchmark based on it. In constructing MRCEval, we employ LLMs as generators for multi-choice sample construction, and judges for challenging samples selection. Extensive studies demonstrate that MRC is still a challenging task for almost all LLMs, especially in relation or event facts understanding and context-faithful. We aim for this work to inspire further advancements in natural language understanding of LLMs.

\section*{Limitations}
MRCEval is an automated construction comprehensive benchmark, it covers a great number of data from other datasets. While we have taken various factors into account, there are a few limitations. First, since we're building on existing datasets, we do not perform refined manual de-noising, but we did perform automated quality detection filtering on the origin data. Secondly, the process of parsing the answers is not completely rigorous. On the one hand, models will output some non-standard responses to a small number of instances, on the other hand, the business models will refuse to answer some questions due to security, ethics and other factors. We do our best to parse the answers from all the responses, but inevitably a small percentage of the sample fails to parse the answers. After analysis, we found that it only accounted for a small part, so it would not have a great impact on the experimental results.

\section*
{Ethical Considerations}
We address several potential ethical considerations in relation to this work:
(1) \textbf{Intellectual property}. This work utilizes several widely adopted MRC datasets, and we fully adhere to their respective licensing agreements. MRCEval will be shared under the CC BY-SA 4.0 license.
(2) \textbf{Intended Use and risk mitigation}. The purpose of this work is to present MRCEval, a benchmark designed to evaluate the capabilities of LLMs on MRC tasks. During the sample selection process, we performed sensitive information filtering on the samples that were rejected by GPT-4o-mini. While we cannot completely rule out the possibility of omissions, we trust in the sensitive information filtering capabilities of GPT-4o-mini. 
(3) \textbf{AI assistance}. GPT-4o was employed to assist in verifying the grammar of the writing.

\bibliography{anthology,custom}

\appendix
\clearpage
\section*{Appendices}

\section{Data Collection}

\subsection{Data Distribution}
\begin{figure}
\centering
\includegraphics[width=\linewidth]{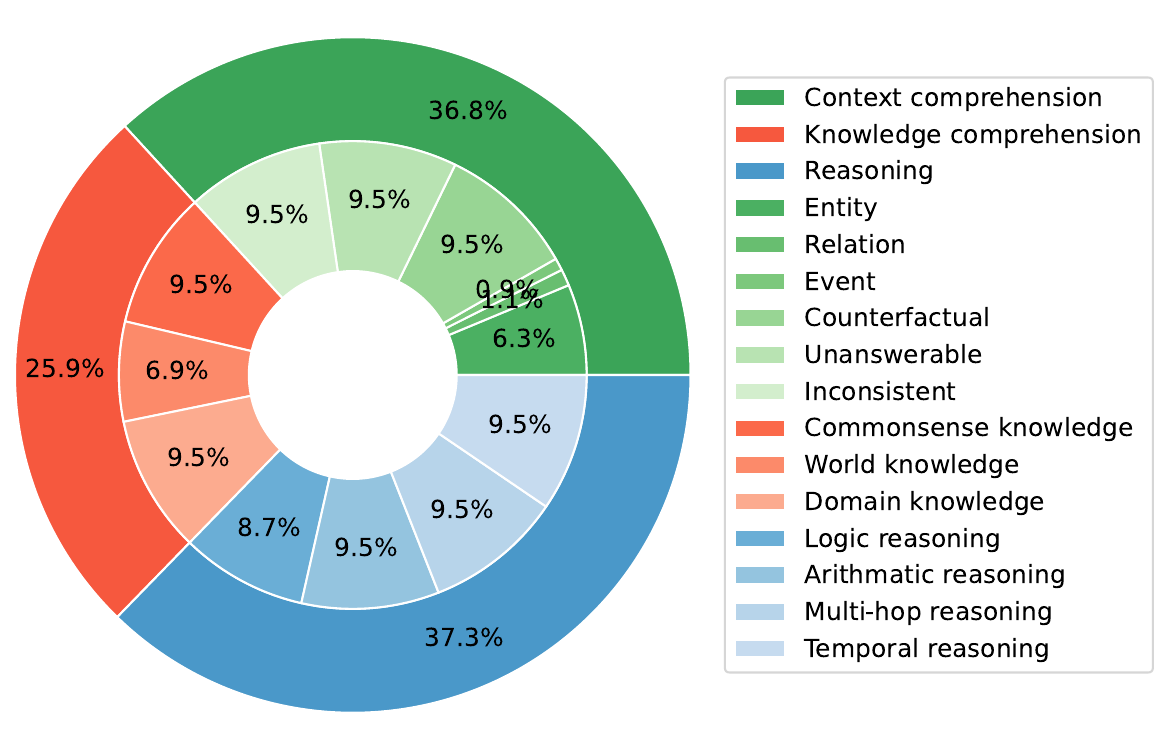}
\caption{Proportion of data in the MRCEval.}
\label{figure:proportion}
\end{figure}

Figure \ref{figure:proportion} shows the distribution of 2,103 instances in MRCEval across three main tasks and 13 sub-tasks. To ensure the data distribution is as uniform as possible, we randomly sample 200 instances for each subtask, except in certain sections where the number of filtered samples is less than 200.

\subsection{Source Datasets}
\paragraph{SQuAD 1.0} \citep{rajpurkar-etal-2016-squad}
(Stanford Question Answering Dataset) is a widely used human-annotated dataset for training and evaluating MRC models, mainly focusing on extracting precise answers from a given passage. It has over 500 articles from Wikipedia, covering diverse topics and more than 100,000+ human-annotated question/answer pairs.
\paragraph{DocRED} \citep{yao-etal-2019-docred}
(Document-Level Relation Extraction Dataset) is a large-scale dataset designed for document-level relation extraction, requiring models to infer entity relations across multiple sentences. It consists of 5,053 documents from Wikipedia, containing 56,354 relational facts from both human-annotated data and distantly supervised data aligned with Wikidata.
\paragraph{MAVEN-Arg} \citep{wang-etal-2024-maven} (MAssive eVENt detection dataset) is a comprehensive and large-scale event understanding dataset, including event detection, event arguments extraction and event relation extraction. It covers 162 event types and 612 argument roles with expert annotation. 
\paragraph{FaithEval} \citep{ming2024faitheval} is a novel benchmark tailored to evaluate the faithfulness of LLMs in contextual scenarios across three diverse tasks: unanswerable, inconsistent, and counterfactual contexts, comprising 4.9K high-quality problems in total. It employs both LLM-based data generation and human validation.
\paragraph{COSMOSQA} \citep{huang-etal-2019-cosmos} is a large-scale dataset of 35,600 multi-choice format problems that require commonsense-based reading comprehension. It focuses on reading between the lines over a diverse collection of people’s everyday narratives, which requires reasoning beyond the exact text spans in the context.
\paragraph{KoRC} \citep{yao-etal-2023-korc} (Knowledge oriented Reading Comprehension) focuses on deep text understanding, which requires the connections between a given document and prior knowledge beyond its text. It has broad knowledge coverage and a flexible answer format. KoRC is constructed based on reasoning chains that weave together documents and a background knowledge base.
\paragraph{PubMedQA} \citep{jin-etal-2019-pubmedqa} is a biomedical question-answering dataset collected from PubMed abstracts. The task of PubMedQA is to answer research questions with yes/no/maybe using the corresponding abstracts. PubMedQA is a well-structured and valuable dataset for advancing AI-driven question-answering in the biomedical field, particularly for applications requiring evidence-based reasoning.
\paragraph{LogicBench} \citep{parmar-etal-2024-logicbench} towards the systematic evaluation of logical reasoning ability of LLMs, which comprehensively evaluates the logical reasoning ability of LLMs on 25 different reasoning patterns spanning over propositional, first-order, and non-monotonic logics. 
\paragraph{DROP} \citep{dua-etal-2019-drop} (Discrete Reasoning Over Paragraphs) requires models to perform discrete reasoning over passages, including numerical operations such as addition, subtraction, and counting, as well as logical reasoning to answer questions correctly. It has crowdsourced and adversarially created 96k questions, which requires a much more comprehensive understanding of the content of paragraphs.
\paragraph{MoreHopQA} \citep{schnitzler2024morehopqa} is a new multi-hop dataset, which shifts from extractive to generative answers. It is created by utilizing three
 existing multi-hop datasets: HotpotQA \citep{yang-etal-2018-hotpotqa}, 2WikiMultihopQA \citep{ho-etal-2020-constructing}, and MuSiQue \citep{trivedi-etal-2022-musique}. It enhances the existing multi-hop questions by adding another layer of questioning that involves one, two, or all three of the following types of reasoning: commonsense, arithmetic, and symbolic.
\paragraph{MenatQA} \citep{wei-etal-2023-menatqa} (Multiple Sensitive Factors
 Time QA) is a dataset for testing the temporal comprehension and reasoning abilities of LLMs. It encompasses three temporal factors (scope factor, order factor, and counterfactual factor) with a total of 2,853 samples.

\section{Benchmark Construction Details}

\subsection{Data Preprocessing.}
To ensure data quality, we first perform sample filtering on the original dataset. For SQuAD and DROP, we filter out samples in the validation set where the answers were inconsistent for denoising. For FaithEval, we find it contains a certain number of problematic samples, so we retain only the data from the NewsQA, RACE, TextbookQA, and HotpotQA sections in FaithEval to ensure the overall sample quality. For PubMedQA, we use their 1k expert-annotated data, and for LogicBench, we use the MCQA section with 500 instances, As for MoreHopQA, we use their 1,118 samples with human verification.

\subsection{Multi-choice Samples Construction}
\begin{figure*}
\centering
\includegraphics[scale=0.8]{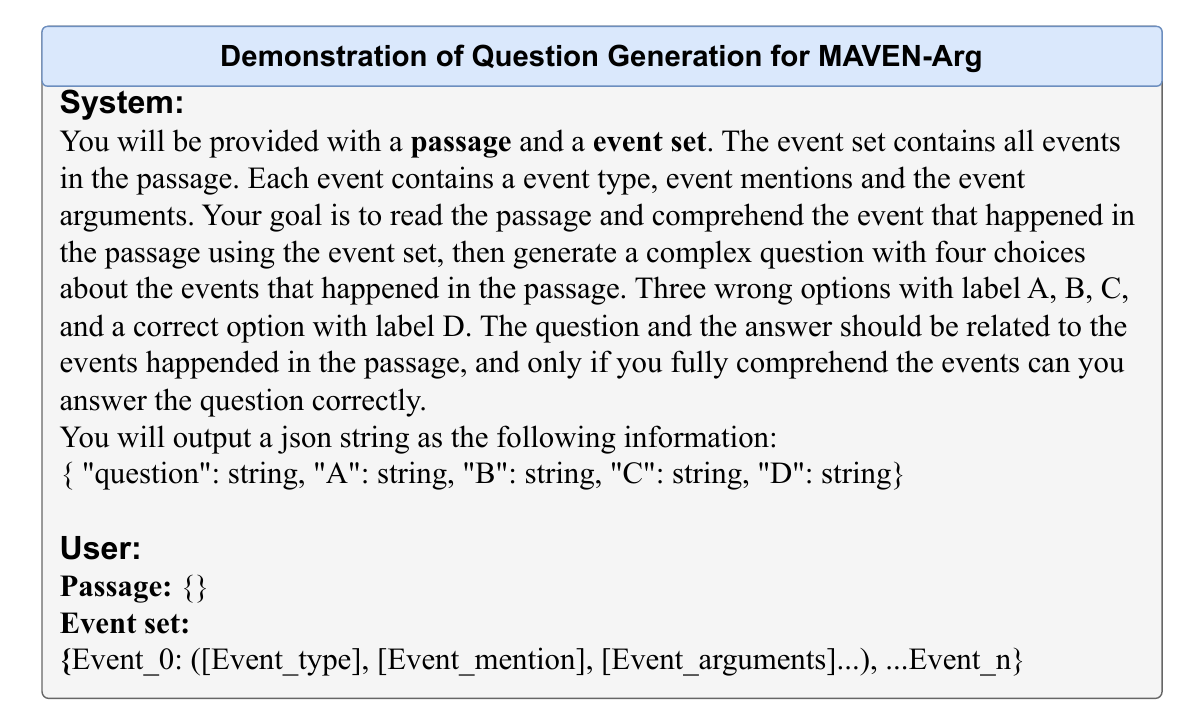}
\caption{Demonstration of question generation for MAVEN-Arg.}
\label{figure:genPrompt_MAVENArg}
\end{figure*}

\begin{figure*}
\centering
\includegraphics[scale=0.8]{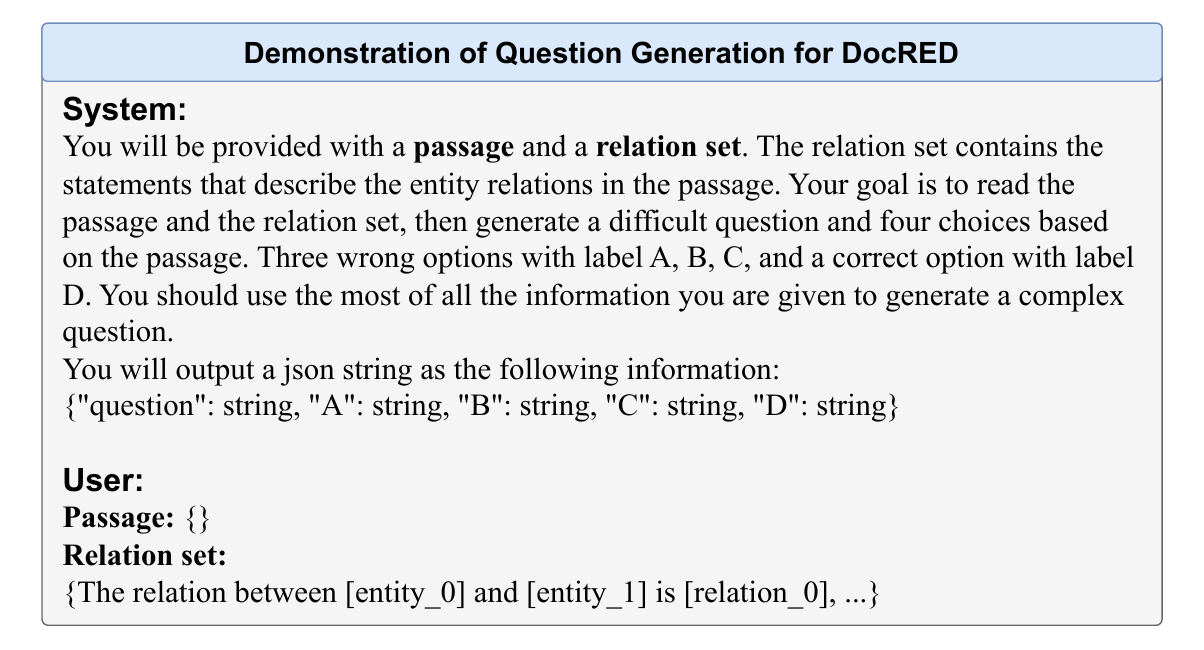}
\caption{Demonstration of question generation for DocRED.}
\label{figure:genPrompt_DocRED}
\end{figure*}

\begin{figure*}
\centering
\includegraphics[scale=0.8]{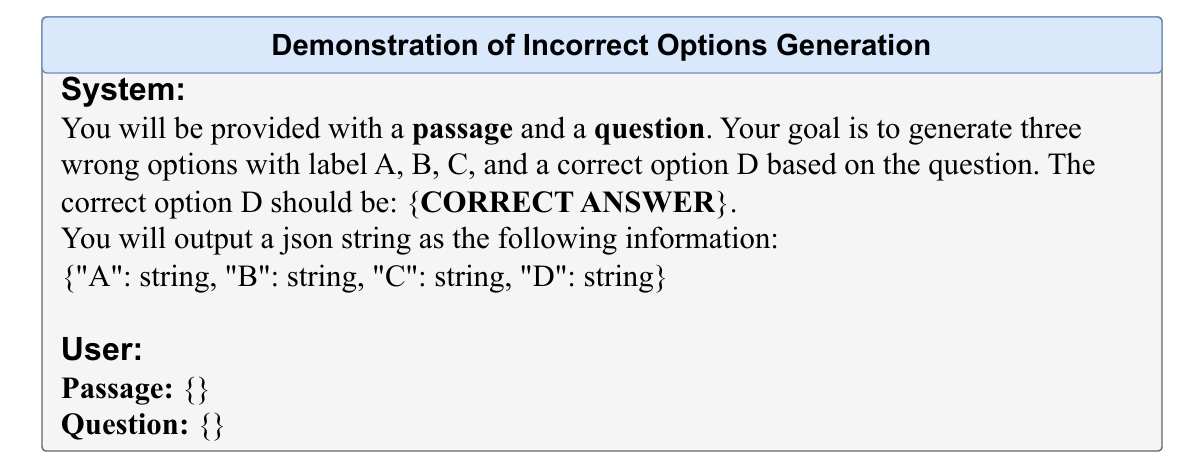}
\caption{Demonstration of incorrect options generation.}
\label{figure:genPrompt_options}
\end{figure*}

To construct a multi-choice benchmark that is convenient for evaluation, we need to generate options or questions for some datasets, as most of them are originally in a QA format.
\paragraph{Questions generation.} For DocRED and MAVEN-Arg, we need to generate one question and four options for each passage. We use GPT-4o as the generator. As Figure \ref{figure:genPrompt_MAVENArg} and \ref{figure:genPrompt_DocRED}, we prompt it to generate facts-related questions and choices with the relation or event information in the datasets.
\paragraph{Options construction.}
For the QA format datasets, we need to construct three incorrect options. For SQuAD 1.0, unanswerable part of FaithEval, KoRC, DROP, and MenatQA, we prompt GPT-4o to generate three incorrect options, as Figure \ref{figure:genPrompt_options}. For others, we use the automated method to construct incorrect options. For inconsistent part of FaithEval, since it has two correct options, we add another two options "Neither A nor B" and "Both A nor B". For PubMedQA, we add "None of the above" option to form four choices. For MoreHopQA, we use the answers of the previous two hops as incorrect choices and add another "None of the above" option. After constructing the four choices, we shuffled their order to mitigate LLMs' selection bias \citep{zheng2024large}.

\subsection{Data Selection}
\begin{figure*}
\centering
\includegraphics[scale=0.55]{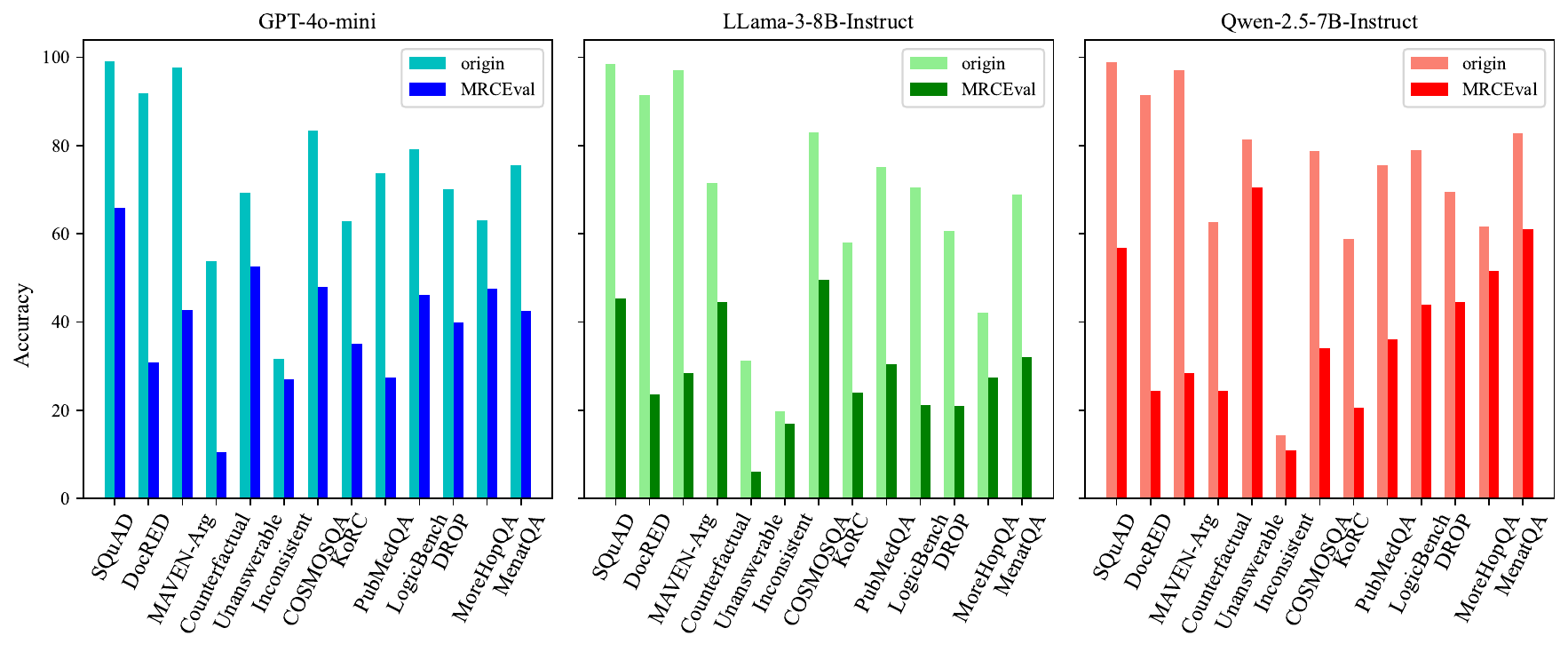}
\caption{Results comparison with the original datasets.}
\label{figure:comparison}
\end{figure*}

\begin{figure*}
\centering
\includegraphics[scale=0.8]{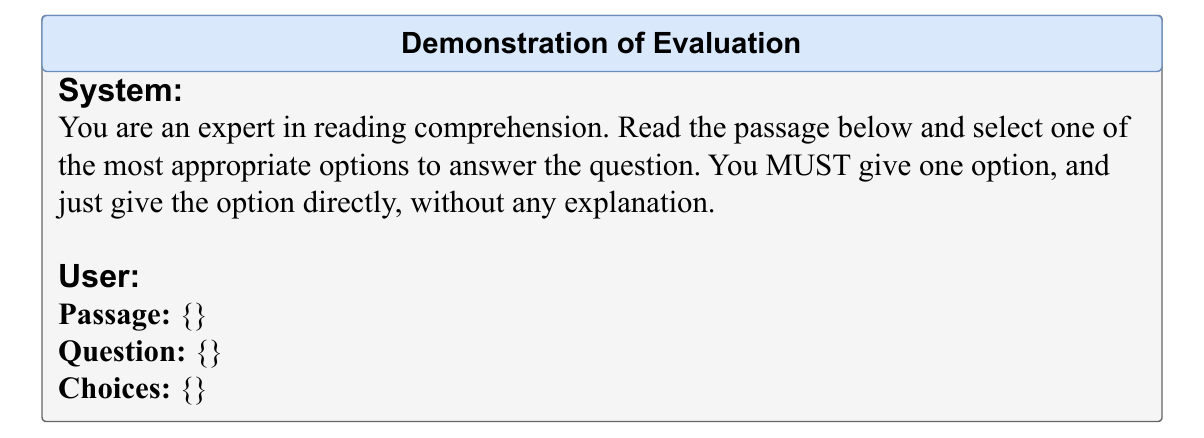}
\caption{Demonstration of Evaluation.}
\label{figure:genPrompt_evaluation}
\end{figure*}

After the construction of multi-choices samples, we employ light-weight LLMs as judges including open-source models LLama-3-8B-Instruct, Qwen-2.5-7B-Instruct and closed-source model GPT-4o-mini. As Figure \ref{figure:genPrompt_evaluation}, we prompt these judges on all the samples, then we get the samples where at least one model gives the incorrect answer. During this process, we find that nearly all models have some instances of refusing to answer or not in the given format, we filter out these samples. Figure \ref{figure:comparison} indicates the comparison with the origin. 

\subsection{Denoising}
Since we select samples misjudged by the models, the filtered samples may contain noise. We conduct noise analysis through sampling. Specifically, we set six powerful LLMs as judges, and sample the instances of each sub-task into three categories: instances that are predicted correctly/incorrectly by all judges, and instances that some judges mispredict. Then we analyze the noise across all sub-tasks, we find that there was little or no noise in most subtasks, except for relation and event facts understanding and world knowledge tasks, which have some noise data. Then we eliminate most noise in these tasks.

\section{Experiment Details}
\paragraph{Experimental setup.} 
We download the open-source models from ModelScope \citep{modelscope} and access closed-source models by calling their official API. Evaluation frameworks are under ModelScope and Hugging Face \citep{wolf2019huggingface}, using 8 NVIDIA RTX 3090 for model inference. 
\paragraph{Model sizes.}
We evaluate a total of 31 popular and latest models, including open-source and proprietary models. A summary of model sizes from different model families is shown in Table \ref{table:model_size}. Due to the high cost, we do not evaluate some expensive but excellent commercial models, including o1, o1-mini with high reasoning effort, o3-mini with high reasoning effort, DeepSeek R1, and Gemini-2.0-flash-thinking.
\paragraph{Expenses.}
The cost of calling the API during the evaluation is approximately \$800-1,000 USD. On top of that, all the experiments take close to a month of GPU time.

\begin{table}
\centering
\resizebox{0.45\textwidth}{!}{
\begin{tabular}{ll}
\toprule
\textbf{Model Name} & \textbf{Size}  \\
\midrule
\multicolumn{2}{c}{\textbf{LLama Family}\citep{dubey2024llama}} \\
\midrule
LLama-3-8B-Instruct & 8B \\
LLama-3.1-8B-Instruct & 8B \\
LLama-3.3-70B-Instruct & 70B \\
\midrule
\multicolumn{2}{c}{\textbf{Qwen Family}\citep{yang2024qwen2}} \\
Qwen-2.5-7B-Instruct & 7B \\
Qwen-2.5-14B-Instruct & 14B \\
Qwen-max-2025-01-25 & unknown \\
\midrule
\multicolumn{2}{c}{\textbf{Mistral Family}\citep{jiang2023mistral}} \\
Mistral-7B-Instruct-v0.3 & 7B \\
Mistral-Nemo-Instruct-2407 & 12B \\
Mistral-8x7B-Instruct-v0.1 & 47B \\
Mistral-large  & unknown \\
\midrule
\multicolumn{2}{c}{\textbf{Gemma Family}\cite{team2024gemma}} \\
Gemma-2-9B-it & 9B \\
Gemma-2-27B-it & 27B \\
\midrule
\multicolumn{2}{c}{\textbf{Phi Family}\citep{abdin2024phi}} \\
Phi-3-mini-4k-Instruct & 3.8B \\
Phi-3-medium-4k-Instruct & 14B \\
Phi-4 & 14B \\
\midrule
\multicolumn{2}{c}{\textbf{Cohere}\citep{command-r}} \\
Command-R-7B-12-2024  & 7B \\
\midrule
\multicolumn{2}{c}{\textbf{DeepSeek}\citep{guo2025deepseek}} \\
DeepSeek-R1-Distll-LLama-8B & 8B \\
DeepSeek-R1-Distll-Qwen-14B& 14B \\
DDeepSeek-v3    & 671B \\
\midrule
\multicolumn{2}{c}{\textbf{OpenAI}} \\
GPT-3.5-turbo & unknown \\
GPT-4-turbo & unknown \\
GPT-4o & unknown \\
GPT-4o-mini  & unknown \\
o1-mini (low reasoning effort) & unknown \\
o3-mini (low reasoning effort) & unknown \\
\midrule
\multicolumn{2}{c}{\textbf{Gemini}} \\
Gemini-1.5-flash & unknown \\
Gemini-1.5-pro & unknown \\
Gemini-2.0-flash & unknown \\
Gemini-2.0-flash-lite-preview-02-05 & unknown \\
\midrule
\multicolumn{2}{c}{\textbf{Anthropic}} \\
Claude-3.5-haiku-20241022 & unknown \\
Claude-3.5-sonnet-20241022 & unknown \\
\bottomrule
\end{tabular}}
\caption{Model size across different model families.}
\label{table:model_size}
\end{table}

\section{Error Study}
\begin{figure}
\centering
\includegraphics[scale=0.45]{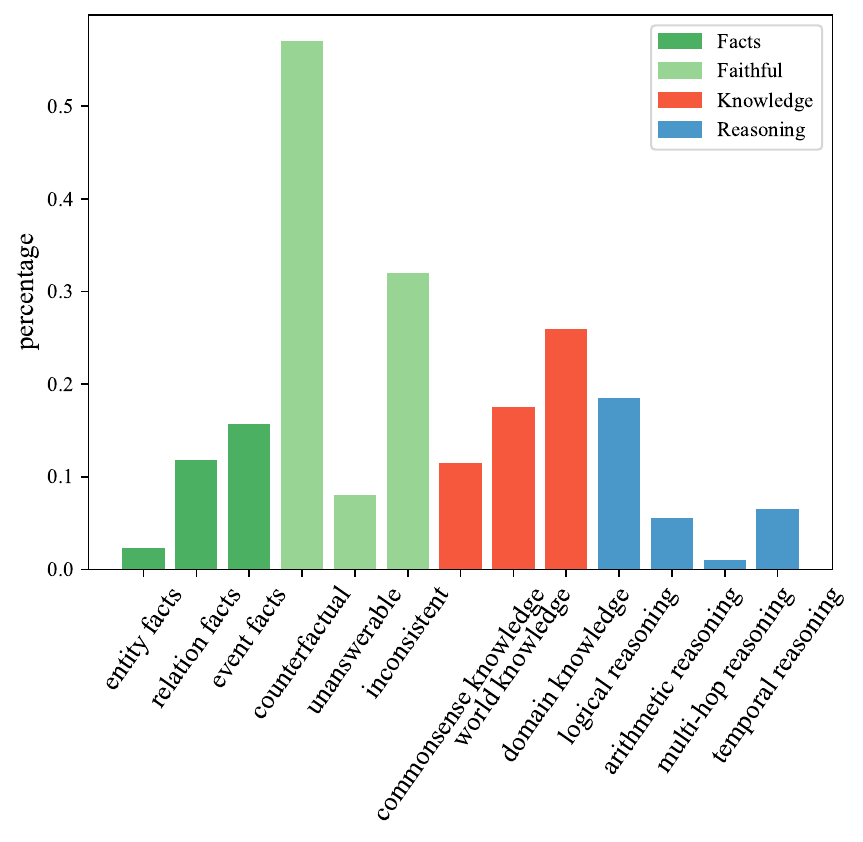}
\caption{The proportion of misjudged samples by all models.}
\label{figure:error_study}
\end{figure}

\begin{figure*}
\centering
\includegraphics[scale=0.8]{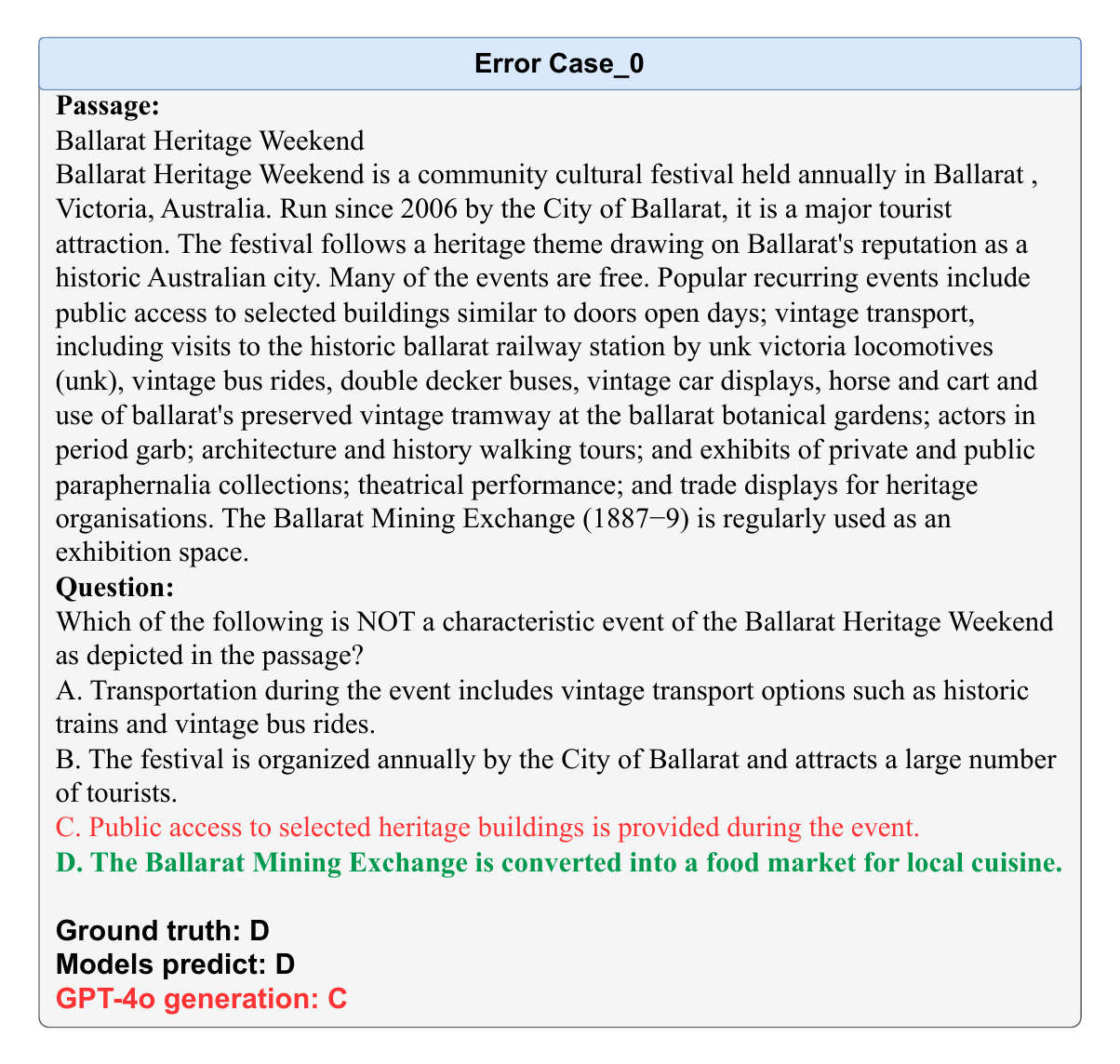}
\caption{Error case when generating by GPT-4o.}
\label{figure:error_MAVEN}
\end{figure*}

\begin{figure*}
\centering
\includegraphics[scale=0.8]{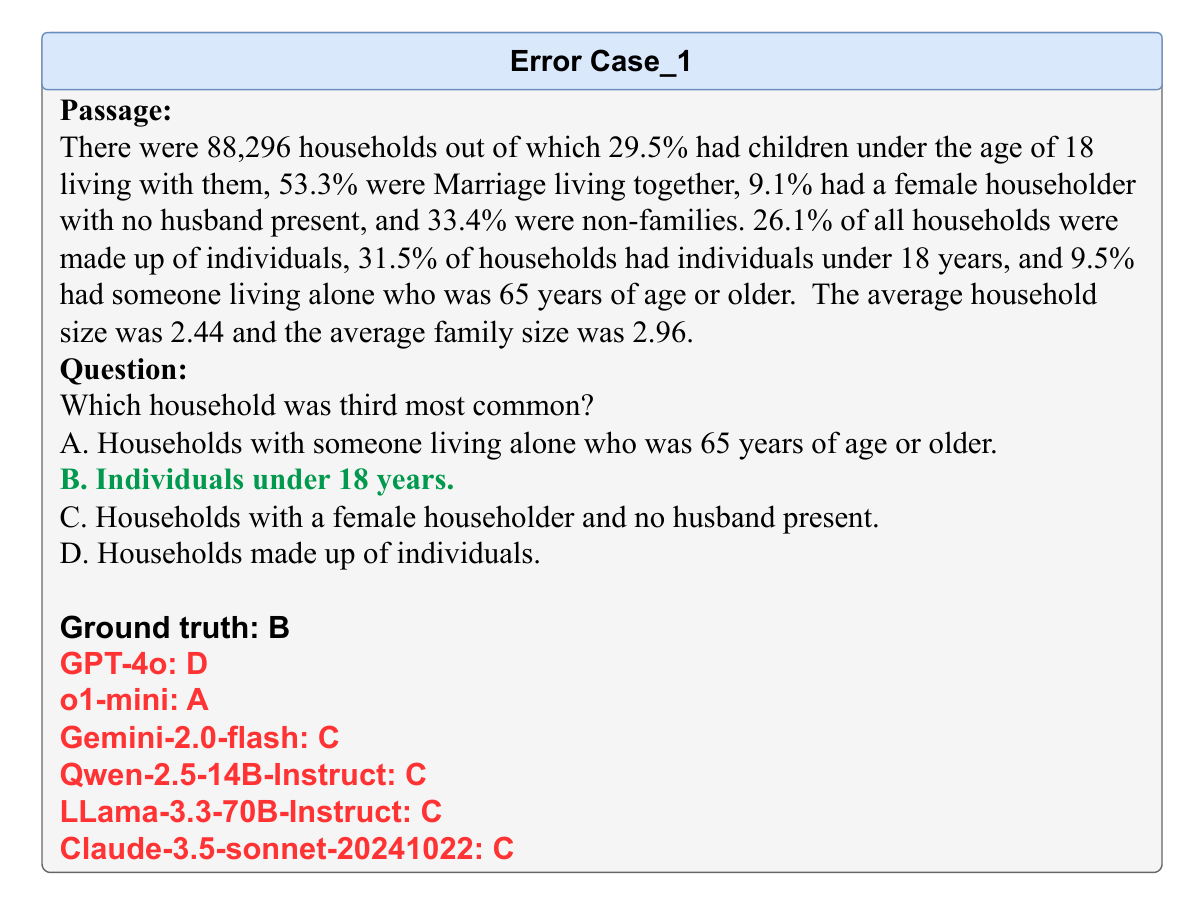}
\caption{Error case from DROP.}
\label{figure:error_DROP}
\end{figure*}

\begin{figure*}
\centering
\includegraphics[scale=0.8]{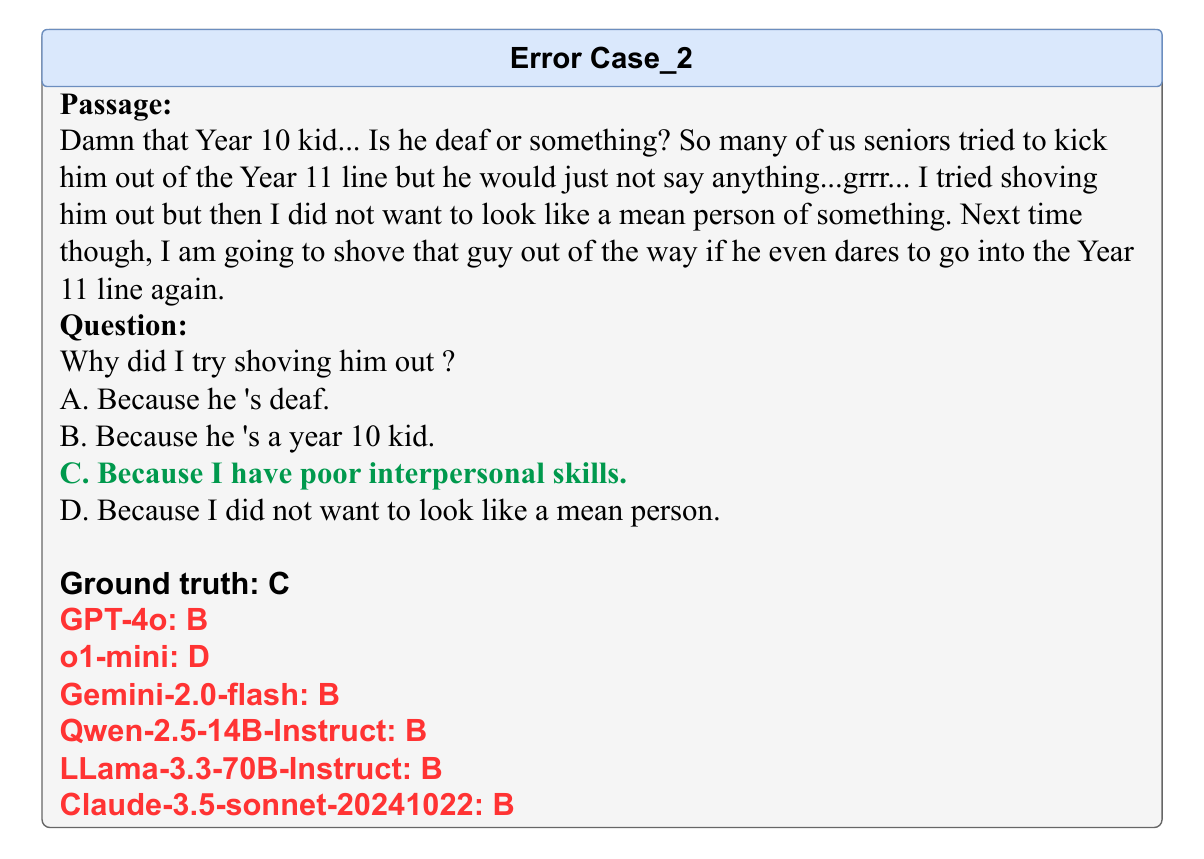}
\caption{Error case from COSMOS.}
\label{figure:error_COSMOS}
\end{figure*}

\begin{figure*}
\centering
\includegraphics[scale=0.8]{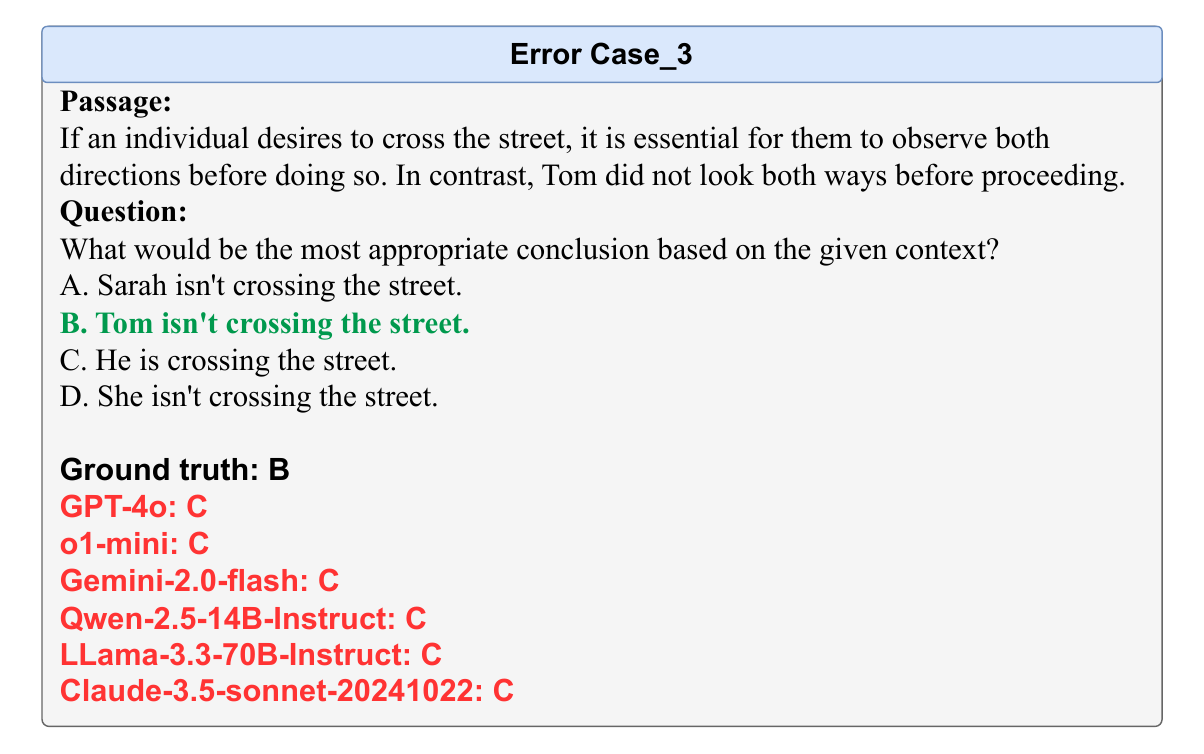}
\caption{Error case from LogicBench.}
\label{figure:error_LogicBench}
\end{figure*}

As Figure \ref{figure:error_study}, the results of the proportion of samples for each sub-task in which all the models (LLama-3.3-70B-Instruct, Qwen-2.5-14B-Instruct, Gemini-2.0-flash, GPT-4o, o1-mini and Claude-3.5-sonnet-20241022) predict incorrectly. 
Apart from this, we sample and analyze the cases from the samples where all these models are predicted incorrectly.
\paragraph{Samples generation noise.} We sample some cases in MAVEN-Arg where questions and choices are generated by GPT-4o. As Figure \ref{figure:error_MAVEN}, we find that not all models predict incorrectly, but GPT-4o generates incorrect answers. The reason may be GPT-4o's loose instruction following where it knows the answer but does not put the correct answer on the appointed label. Apart from this, there are other noises like wrong questions, duplicate options, and multiple correct options when generating with GPT-4o.
\paragraph{Models prediction errors.} Others are common errors as we clarify, like arithmetic errors as Figure \ref{figure:error_DROP}, commonsense errors as Figure \ref{figure:error_COSMOS} and logic errors as Figure \ref{figure:error_LogicBench}. From these examples, we can also see that the model relies more on common situations and context to infer what may happen, rather than fully understanding the given text and reasoning.

\end{document}